\documentclass[fleqn,10pt]{wlscirep}
\usepackage{booktabs}       % professional-quality tables
\usepackage{graphicx}
\usepackage{subcaption}
\usepackage{color, comment}
\usepackage{algorithm}
\usepackage{algorithmic}
\usepackage{footnote}
\usepackage{tablefootnote}
\usepackage{amsmath}
\usepackage{amssymb}
\usepackage{float}
\title{Deep Reinforcement Learning for Dynamic Treatment Regimes on Medical Registry Data}

\author[1,+]{Ning Liu}
\author[2,*+]{Ying Liu}
\author[2]{Brent Logan}
\author[1]{Zhiyuan Xu}
\author[1]{Jian Tang}
\author[1]{Yanzhi Wang}
\affil[1]{Department of Electrical Engineering and Computer Science, Syracuse University, Syracuse, NY 13244, USA}
\affil[2]{Division of Biostatistics, Medical College of Wisconsin, Milwaukee, WI 53226, USA}

\affil[*]{corresponding author email: summeryingl@gmail.com}

\affil[+]{these authors contributed equally to this work}


\begin{abstract}
This paper presents the first deep reinforcement learning (DRL) framework to estimate the optimal Dynamic Treatment Regimes from observational medical data. This framework is more flexible and adaptive for high dimensional action and state spaces than existing reinforcement learning methods to model real-life complexity in heterogeneous disease progression and treatment choices, with the goal of providing doctor and patients the data-driven personalized decision recommendations.
The proposed DRL framework comprises (i) a supervised learning step to predict the most possible expert actions, and (ii) a deep reinforcement learning step to estimate the long-term value function of Dynamic Treatment Regimes. Both steps depend on deep neural networks.
 As a key motivational example, we have implemented the proposed framework on a data set from the Center for International Bone Marrow Transplant Research (CIBMTR) registry database, focusing on the sequence of prevention and treatments for acute and chronic graft versus host disease after transplantation.
In the experimental results, we have demonstrated promising accuracy in predicting human experts' decisions, as well as the high expected reward function in the DRL-based dynamic treatment regimes.

\end{abstract}

\begin{document}

\flushbottom
\maketitle
%% This is `naturemag-doi.bst` v1.3, modified from naturemag.bst
\thispagestyle{empty}

\section*{Introduction}
Medical treatments often compose a sequence of intervention decisions that are made adaptive to the time-varying clinical status and conditions of a patient, which are coined as \emph{Dynamic Treatment Regimes} (DTRs \cite{Lavori2000}). ``How can we optimize the sequence of specific treatments for specific patients?'' is a central question of \emph{precision medicine}. More specifically, the scientific question our paper focuses on is the determination of the optimal DTRs to maximize the long-term clinical outcome.

When the straightforward rule-based treatment guidelines are difficult to be established, the statistical learning method provides a data-driven tool to explore and examine the best strategies. These data driven approaches leverage on the technology advances to collect the increasingly abundant medical data (e.g., clinical assessments, genomic data, electronic health records) from each individual patient to meet the promise of individualized treatment and health care.

The problem of identifying the optimal DTRs that maximize the long-term clinical outcome using \emph{reinforcement learning} \cite{sutton1998reinforcement} has received much attention in the statistics community \cite{moodie2007,lavori2004,murphy2003,robins2004,robust2012,zhao2009,murphy2007,zhao2014,liu2016robust}.
The existing DTR methods are proposed on \emph{Sequential Multiple Assignment Randomized Trial} (SMART) \cite{Murphy2005}, in which the methods for DTR optimization are limited to clearly defined homogeneous decision stages and low-dimensional action spaces. They are difficult to implement using observational data (such as electronic medical records, registry data), which exhibit much higher degree of heterogeneity in decision stages among patients, and the treatment options (i.e., the action space) are often high-dimensional. The existing methods can only analyze certain simplification of stage and action spaces among the enormous ways. Simplification by human experts might not lead to the optimal DTRs and in many cases there is no clear way of simplification. In addition, the simplification process needs substantial domain knowledge and labor-intensive data mining and feature engineering processes. There is a call for methods to expand DTR methodology from the limited application in SMART studies to broader, flexible, and practical applications using registry and cohort data.

To make reinforcement learning accessible for more general DTR problems using observational datasets, we need a new framework which (i) automatically extracts and organizes the discriminative information from the data, and (ii) can explore high-dimensional action and state spaces and make personalized treatment recommendations. \emph{Deep learning} is a promising new technique to use \emph{representation learning} and save the labor-intensive feature engineering processes.
The effective combination of deep learning (deep neural networks) and reinforcement learning technique, named \emph{Deep Reinforcement Learning} (DRL), is initially invented for intelligent game playing and has later emerged as an effective method to solve complicated control problems with large-scale, high-dimensional state and action spaces \cite{mnih2013playing,mnih2015human,silver2016mastering,wang2017dac,wang2017icdcs,wang2017icc} .
The deep learning and DRL methods are promising to  automatically extract discriminate information among decision stages, patient features, and treatment options. In the work we incorporate the state-of-the-art deep reinforcement learning into the DTR methodology and propose the first (to the best of our knowledge) data-driven framework that is scalable and adaptable to optimizing DTR with high-dimensional treatment options, and heterogeneous decision stages.

%Through effective incorporating deep learning and DRL methods to  automatically extract discriminate information among decision stages, patient features, and treatment options, in the work we propose the first (to the best of our knowledge) data-driven framework that is scalable and adaptable to optimizing DTR with high dimensional treatment options, and heterogeneous decision stages.

To demonstrate the effectiveness of the proposed framework, we implemented it using a concrete example: Graft Versus Host Disease (GVHD) prevention and treatment for Leukemia patients who have undergone allogeneic hematopoietic cell transplantation (AHCT). The long-term longitudinal follow-up for almost all US patients and some international patients who have undergone AHCT make the Center for International Blood and Marrow Transplant Research (CIBMTR) registry database an ideal existing data set to explore the capacity of artificial intelligence in medical decision making. %Our target cohort has 43319 patients, which includes almost the whole target disease population in the U.S. This is an important step towards the next generation of automated and universal medical decision analysis tool which  is less dependent on human domain knowledge and feature engineering. \textcolor{red}{Although the above sentence claimed the unique advantage of this data set for exploring the power of DRL in medical observational data, but we may not want to expose or emphasis the fact that we havenot collected all these data? }

Reference \cite{ruutu2014prophylaxis} points out that GVHD is a major complication of AHCT. Once established, GVHD is difficult to treat. It can be prevented by selected methods, but often at the expense of an increased risk of relapse, rejection or delayed immune reconstitution \cite{bacigalupo1991increased, patterson1986graft}. Hence, no optimal or even satisfactory prevention and treatment methods have been defined.
 Reference \cite{ruutu2014prophylaxis} concluded that the difficulty in composing a standard practice guideline is the lack of solid scientific support for a large portion of procedures used in GVHD prevention and treatment, which calls for further systematic studies to compare different strategies. Such clinical needs for methodological innovations in finding the optimal personalized strategies can be largely resolved in the proposed study.

More specifically, in this paper we develop a data-driven deep reinforcement learning (DRL) framework for the optimal DTR, comprising the prevention and treatment of both acute and chronic GVHDs, as well as the initial conditioning (chemotherapy) after the transplantation. The DRL framework, which deals with heterogenous decision stages (states) and high-dimensional action space, consists of two steps at each decision stage. The first step is to build a deep neural network to predict experts' treatment with a high-dimensional action space. The second step is to estimate the \emph{value function} of DTRs for strategies composed of the top expert actions with highest probability from the first step. The state and action spaces as well as reward function are carefully selected, and effective dimensionality reduction techniques such as the \emph{low variance filter} are utilized to mitigate the shortcoming of limited data in the database. The similar states have similar encoded representations.
In the experimental results, we have demonstrated promising accuracy in predicting human experts' decisions, as well as the high expected reward function in the DRL-based dynamic treatment regimes.

\section*{Results}

In this section, we present results on the deep neural networks' prediction accuracy for expert treatment as well as the performance of the deep reinforcement learning  for optimizing the sequence of treatments. % In this section, we present the extensive data-driven experimental results of prevention and treatment of both acute and chronic GVHDs, as well as the initial conditioning (chemotherapy) after the transplantation.
Experiments are conducted based on the CIBMTR registry with data of 6,021 patients. The initial conditioning (to prevent relapse) and GVHD prophylaxis (to prevent GVHD) were administered right before the transplant, thus they are considered as action at time $t=0$; the treatment of acute GVHD takes place at 100 days and 6 months (180 days); the treatment of chronic GVHD takes place at 6 months, 1 year (365 days), 2 years (730 days), and 4 years. We test DTR within 4 years after transplantation because a large portion of patients' data will be missing after that time, and live patients without relapse can be considered to be cured from the disease.
In the following, we will demonstrate the data-driven experimental results on the first step, i.e., building a deep neural network to predict experts' treatment, and then the second step, i.e., DRL-based framework of value function estimation and making recommendations among treatment options. We adopt separate DNNs for predicting experts' treatment in the first step, and separate DRLs for treatment of acute and chronic GVHDs in the second step. This is because of the limited data size to train an overall large DNN or DRL model. Details of the proposed procedure will be discussed in the next section.

\subsection*{Results on Predicting Experts' Treatment}
First, we demonstrate in Figure 1 the prediction accuracies of the initial conditioning and the initial GVHD prevention (prophylaxis). We use 80\% of the data set as training data and the remaining 20\% for testing data, which is common for the deep learning testings. Please note that we utilize the top-$N$ prediction accuracy, i.e., the prediction is valid as long as the actual treatment action from human experts is among the top $N$ choices suggested by the deep neural network. This top-$N$ accuracy is widely utilized for the image recognition such as the ImageNet contest \cite{deng2009imagenet} and other deep learning testings. We can observe that (i) the top-$N$ accuracy is in general between 75\% and 90\%, which shows the effectiveness of the proposed method; and (ii) the top-$N$ accuracy will increase with the increase of the $N$ value.

\begin{figure}[H]
\centering
\begin{subfigure}{.52\textwidth}
  \centering
  \includegraphics[width=.8\linewidth]{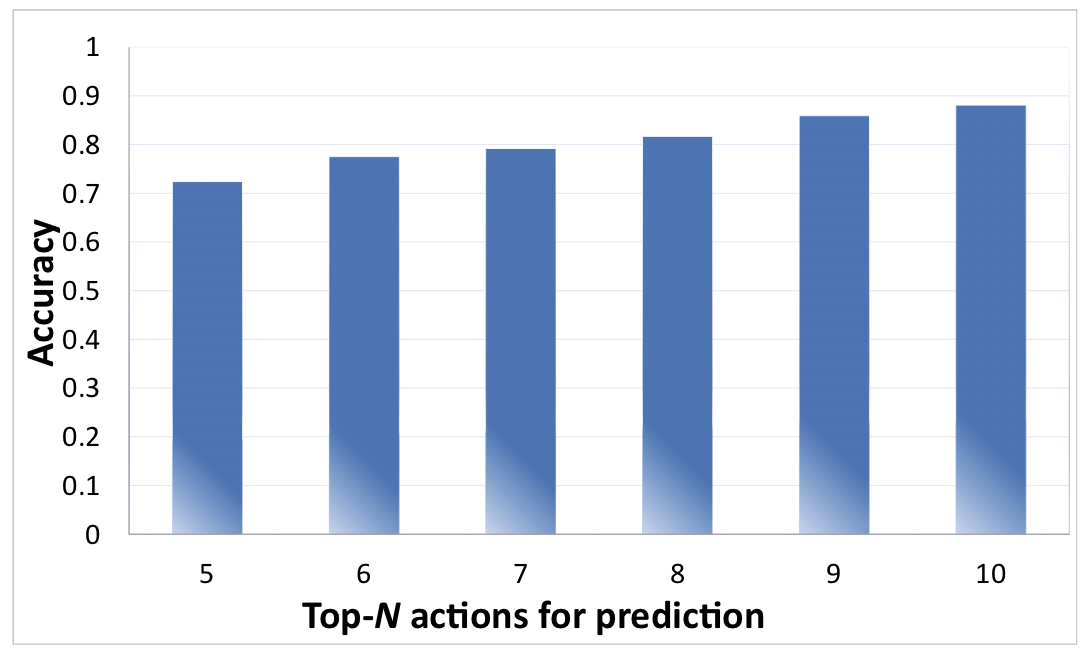}
  \caption{Initial conditioning}
  \label{fig:inital_bsconditioninge}
\end{subfigure}%
\begin{subfigure}{.52\textwidth}
  \centering
  \includegraphics[width=.8\linewidth]{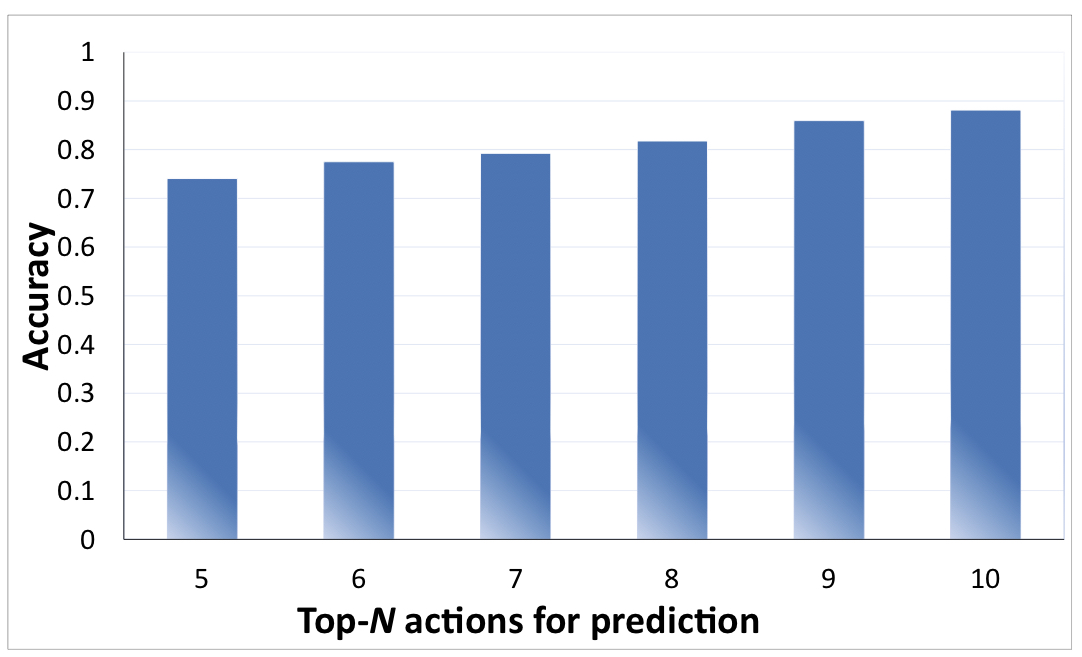}
  \caption{Initial GVHD prophylaxis}
  \label{fig:inital_bsgvhd}
\end{subfigure}
\caption{Accuracies on predicting experts' treatment for initial conditioning and GVHD prophylaxis.}
\label{fig:initial_medical}
\end{figure}

Furthermore, Figure 2 illustrates the top-$N$ prediction accuracy results for acute GVHD treatments at (a) 100 days and (b) 6 months. Figure 3 illustrates the (a) top-7 and (b) top-10 prediction accuracy results for chronic GVHD treatments, at 100 days, 6 months, 1 year, and 2 years. Again we use 80\% of the data set as training data and the remaining 20\% for testing data. From these two figures we can derive the following observations. First, the prediction accuracies are in general higher compared with the initial conditioning and GVHD preventions, because the medication for GVHD treatments seems to be more regular compared with the initial treatments. The prediction accuracies are high enough and this shows a first step towards the ultimate goal of DTR using machine intelligence. Next, for the chronic GVHD treatment, the prediction accuracy will increase when time elapses, i.e., the prediction accuracy at 180 days is higher than that at 100 days, and the accuracy at 1 year will be even higher. The reason is that the patients will become more stable and easy for treatment when chronic GVHD occurs or prolongs at a later time.

\begin{figure}[H]
\centering
\begin{subfigure}{.52\textwidth}
  \centering
  \includegraphics[width=.8\linewidth]{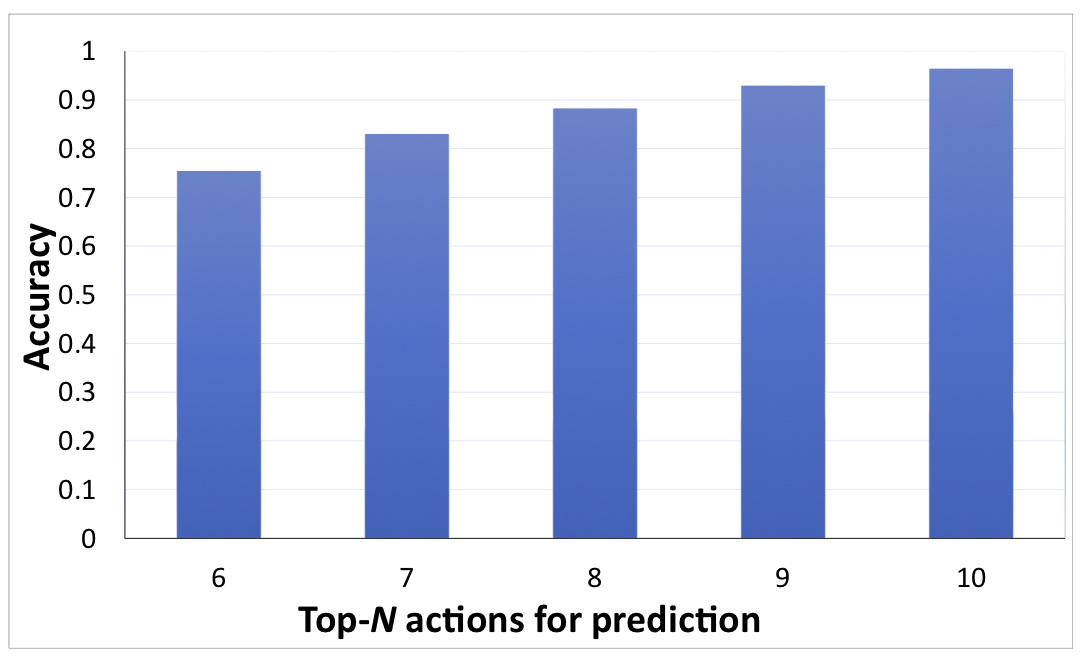}
  \caption{Results at 100 days}
  \label{fig:DRL_Agvhd_6monthacc}
\end{subfigure}%
\begin{subfigure}{.52\textwidth}
  \centering
  \includegraphics[width=.8\linewidth]{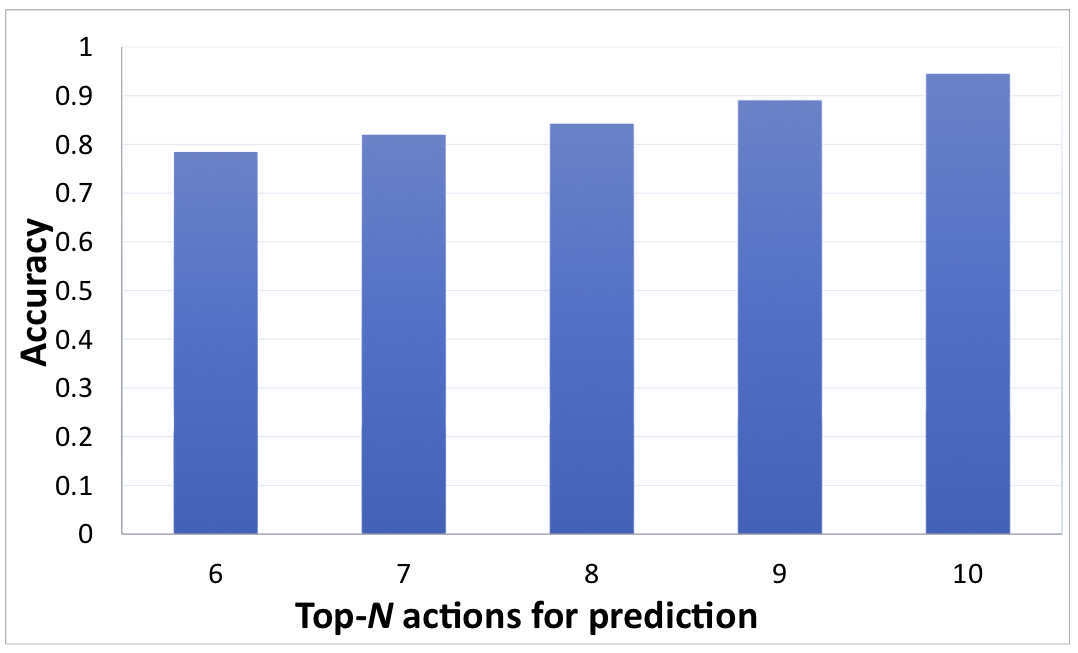}
  \caption{Results at 6 months}
  \label{fig:DRL_Agvhd_1yearacc}
\end{subfigure}
%\begin{subfigure}{.33\textwidth}
%  \centering
%  \includegraphics[width=.9\linewidth]{DRL_Agvhd_reward.jpg}
%  \caption{DRL Agvhd reward}
%  \label{fig:DRL_Agvhd_reward}
%\end{subfigure}
\caption{Accuracies on predicting experts' treatment for acute GVHD.}
\label{fig:DRL_Agvhd}
\end{figure}

\begin{figure}[H]
\centering
\begin{subfigure}{.52\textwidth}
  \centering
  \includegraphics[width=.8\linewidth]{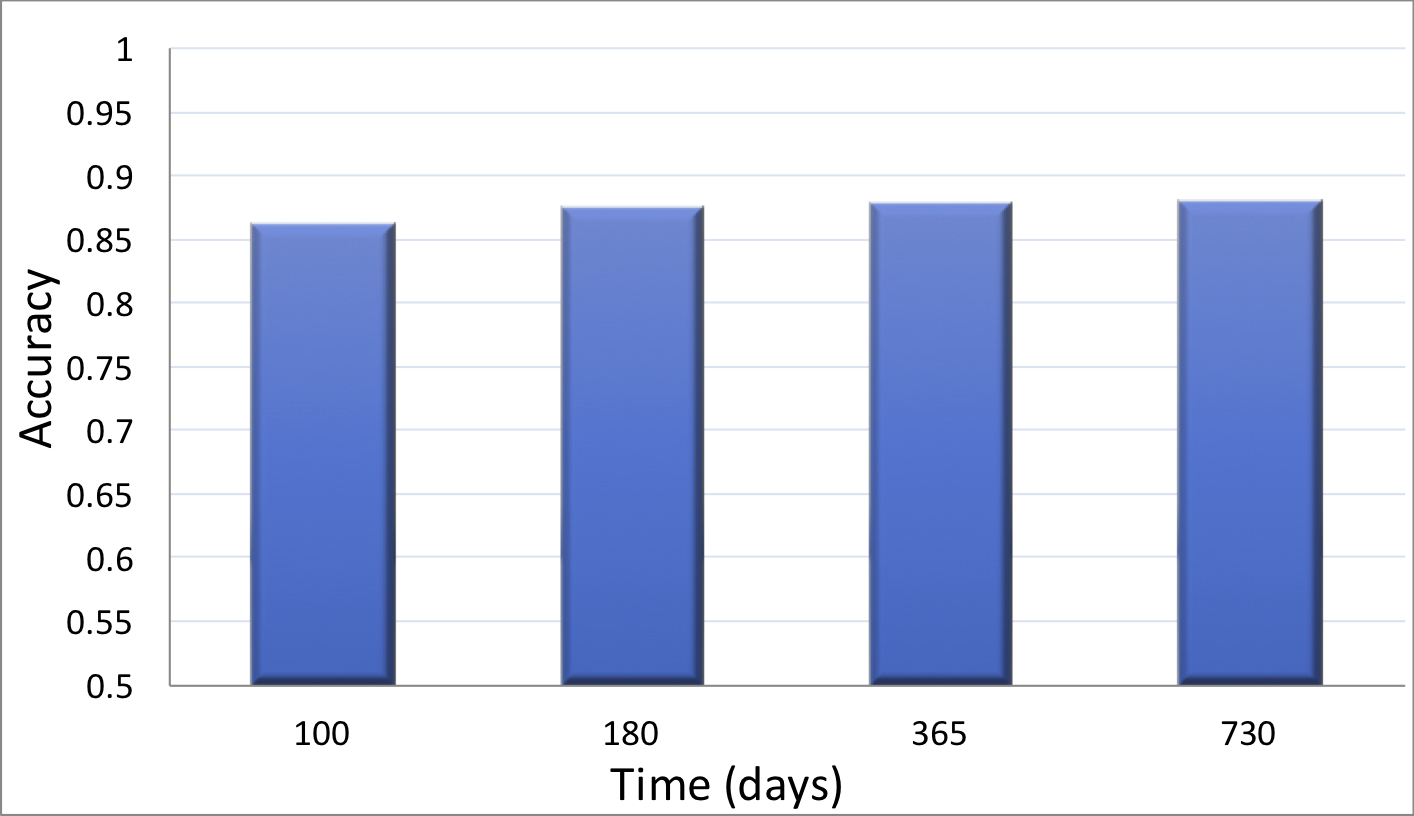}
  \caption{Top-7 accuracy results}
  \label{fig:DRL_cgvhd_top7acc}
\end{subfigure}%
\begin{subfigure}{.52\textwidth}
  \centering
  \includegraphics[width=.8\linewidth]{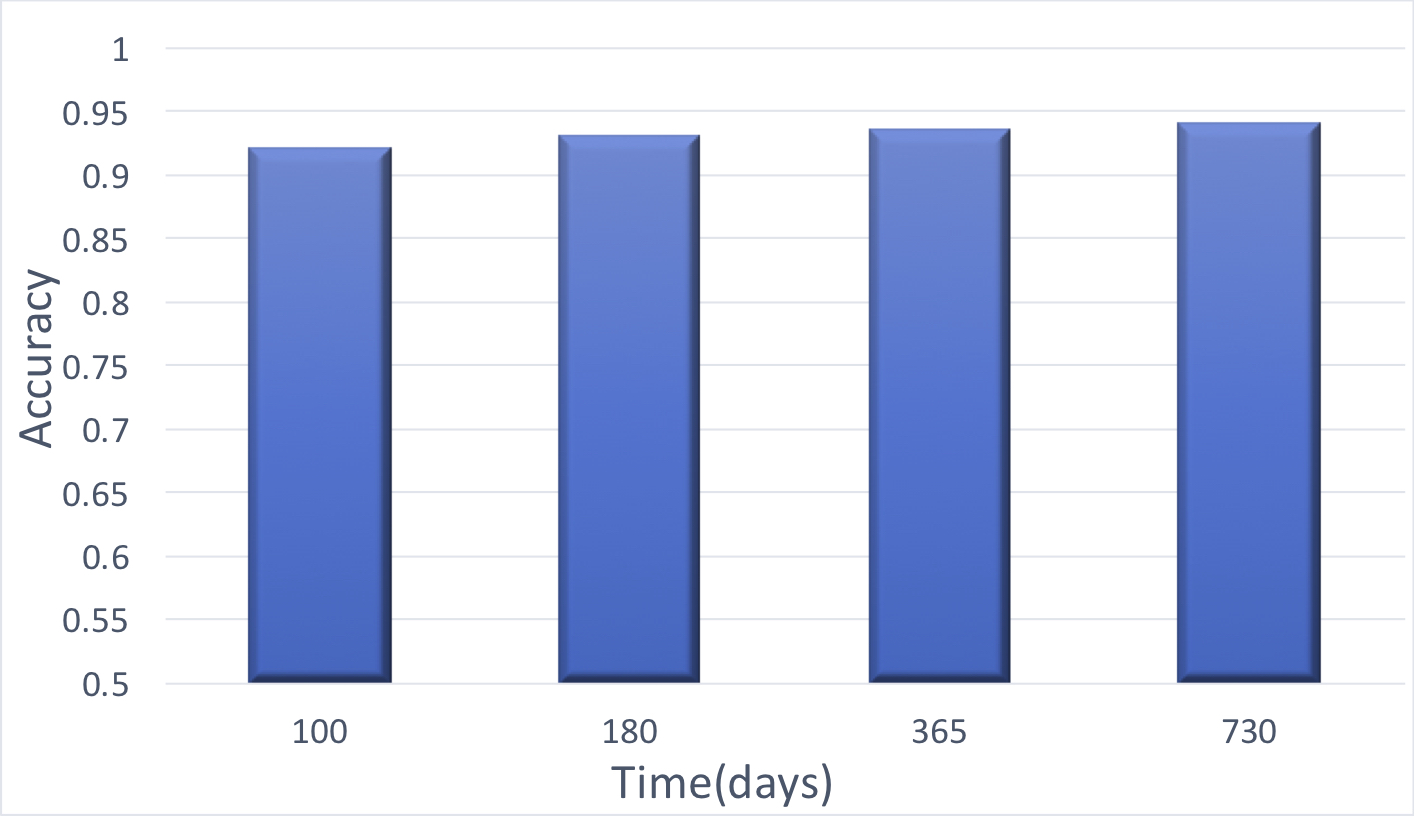}
  \caption{Top-10 accuracy results}
  \label{fig:DRL_cgvhd_top10acc}
\end{subfigure}
%\begin{subfigure}{.33\textwidth}
%  \centering
%  \includegraphics[width=.9\linewidth]{DRL_cgvhd_reward.jpg}
%  \caption{DRL Agvhd reward}
%  \label{fig:DRL_cgvhd_reward}
%\end{subfigure}
\caption{Accuracy results on predicting experts' treatment for chronic GVHD at 100 days, 6 months, 1 year, and 2 years.}
\label{fig:DRL_Cgvhd}
\end{figure}

\subsection*{Results on DRL-based Value Function Estimation and Making Recommendations}

In this section, we provide experimental results on the effectiveness of the DRL-based DTR framework for acute and chronic GVHD treatments, i.e., using DRL for value function estimation and making recommendations. Because of the limited data size to train an overall large DRL model, we build separate DRL models for the treatments of acute and chronic GVHDs. Details about the DRL models are provided in the next section. Again we use 80\% of the data set as training data and the remaining 20\% for testing data.

The reader of the paper will be most interested in how the DRL-based recommendation making would improve the cumulative outcome, i.e., the disease free survival, of the patients.
As a result, we compare between the proposed DRL-based approach with random action selection baseline in terms of the value function. The details of value function are described in the next section (we use the highest value 1 for relapse-free and GVHD-free survival, the lowest value 0 for death, and values in between for other terminal states such as relapse, GVHDs, etc.)
More specifically, the baseline uses the average DRL values of all available actions (excluding the one with the highest expected reward) to mimic a random action selection policy. Using the actual values from he observational data set results in similar baseline performance and will not be shown in this paper.

Figure 4 illustrates the comparison results between the proposed DRL method and baseline for acute GVHD treatment, while Figure 5 shows the comparison results for chronic GVHD treatment.
Despite the limited data, we can still observe that the proposed DRL method outperforms the baseline methods both for acute and chronic GVHD treatments, which illustrates the effectiveness of using the DRL method for making recommendations in DTR.
Also, we can observe that the value function (accumulative rewards) will increase when time elapses.
This observation can also be explained as the expected outcome (e.g., the final relapse-free survival rate) will become higher for a patient if he/she has survived without relapse over a period of time (say 1 year or 2 years).

\begin{figure}[H]
\centering
\includegraphics[width=.65\linewidth]{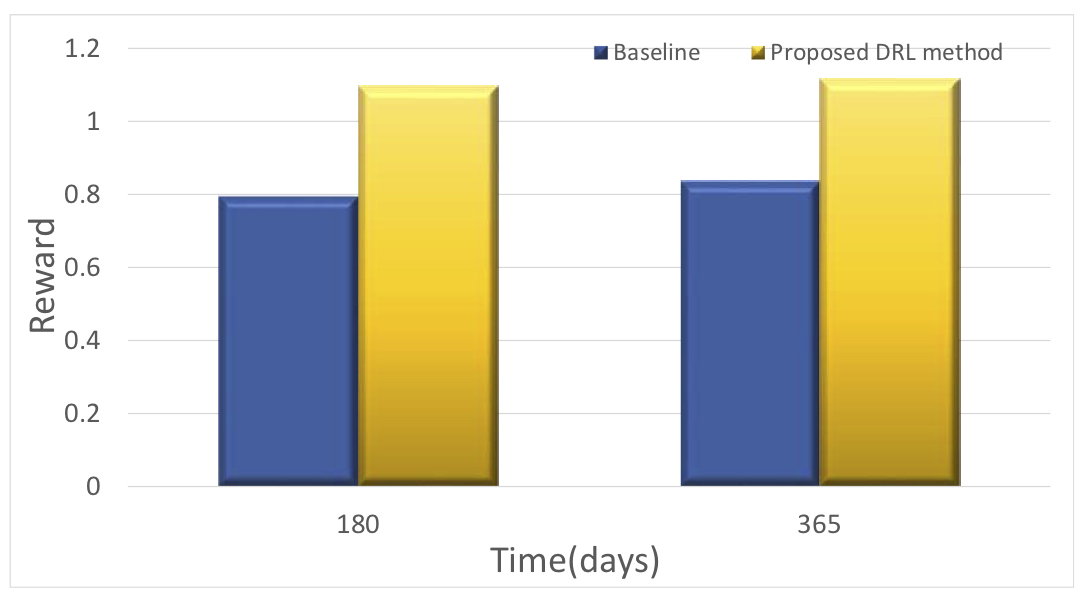}
\caption{Comparison results between the proposed DRL method and the baseline (details in the context) for acute GVHD treatment.}
\label{fig:DRL_Agvhd_reward}
\end{figure}

\begin{figure}[H]
\centering
\includegraphics[width=.65\linewidth]{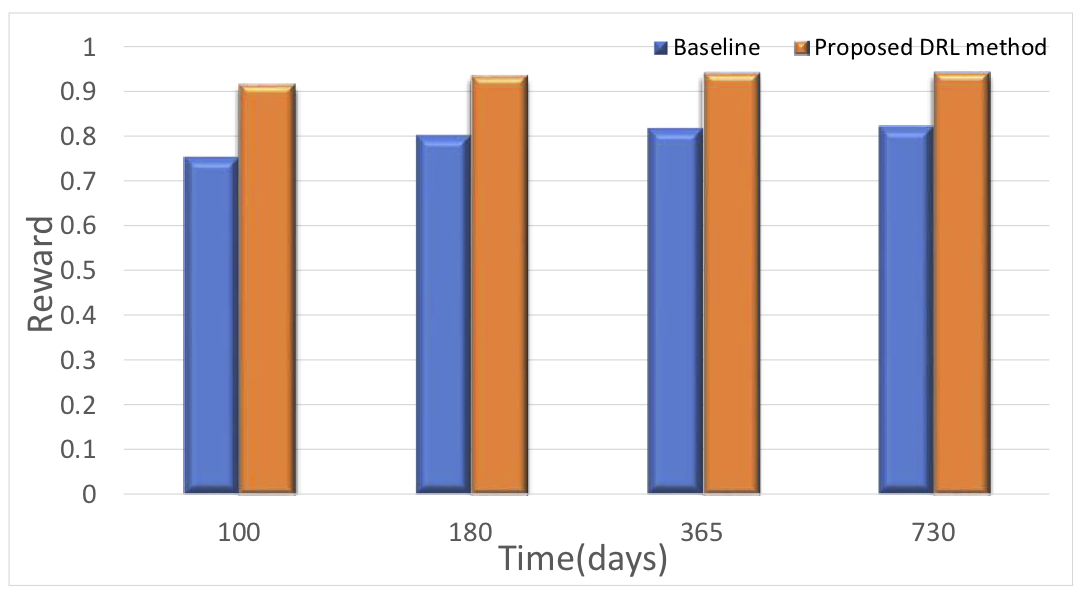}
\caption{Comparison results between the proposed DRL method and the baseline (details in the context) for chronic GVHD treatment.}
\label{fig:DRL_cgvhd_reward}
\end{figure}

\section*{Discussion}

In this work, we present a machine learning strategy from an observational dataset to address the decision making problem of GVHD prevention and treatment. It is of significant interests to incorporate this machine learned rule to facilitate treatment decision making and how to update the decision rules in an online fashion when new data are collected.
There are some current trends in the mobile health field that combines randomized clinical trial with online reinforcement learning through micro-randomized-trials \cite{klasnja2015microrandomized}, where the randomization probability can be adopted in an online manner, in analogy to the exploration techniques in reinforcement learning. The applications can be seen in smoking secession, eating disorder management, and blood glucose management for diabetes patients. However, compared with our motivating example in bone marrow transplant, these existing interventions are easier to perform randomization due to the much fewer number of actions, less profound consequences, fewer treatment options and less complicated feature variables.

 %The more established SMART studies are most commonly conducted in the mental health field, but it has not been conducted in the leukemia field due to much larger resources and more ethical concerns in these fields.

   Nevertheless, in the clinical fields like our motivating example, there are some pressing sequential decision making questions. For example, in the leukemia field, one other example is to decide whether transplant is a beneficial strategy compared to non-transplant, under what condition or time transplant will become a better option, and adapting these decisions to personal features.
   Given the constraints on conducting sequential randomized clinical trials on these questions, it is more practical to start from analyzing the observational data at this point. With the improvement of data collection and machine learning techniques in this field, a data-driven decision support system can provide treatment recommendations for doctors based on supervised learning and reinforcement learning. Furthermore, one can adopt the exploration policy in reinforcement learning for the adaptive treatment recommendations, while the decision is made through doctors and patient's preference.

 The Q-learning has theoretical guarantee to converge to optimal policy only under the assumptions for Markov decision process. However Deep Q-learning does not have theoretical guarantee for convergence to optimal policy even under the Markov decision process because of the sub-optimality of deep neural networks. The disease progression process does not strictly follow Markov process and the four state variables we are considering may not fully capture the patients status. However Q-learning and DQN have demonstrated good performance in a lot of application that Markov (memoriless) property does not hold \cite{liu2017hierarchical,zhu2017target}. For future work we will remedy for this problem with model without Markov assumption (e.g. RNN), taking the history information into account.

\section*{Methods}

In this section we discuss in details the proposed DRL framework for the optimal DTR, comprising the prevention and treatment of both acute and chronic GVHDs, as well as the initial conditioning after transplantation. We first provide a general framework of DRL which can deal with complicated control problems with high-dimensional state spaces, and then the cohort retrieving and data pre-processing, problem formulation, state and action spaces, reward function, and optimization techniques of the proposed DRL framework for precision medicine.

\subsection*{The General DRL Framework for Complicated Control Problems}

The general DRL framework, which can be utilized to solve complicated control problems, consists of two phases: an offline deep neural network (DNN) construction phase and an online deep Q-learning phase\cite{mnih2013playing,mnih2015human,silver2016mastering}. In the offline phase, a DNN is utilized to derive the correlation between each state-action pair $(s,a)$ of the system under control and the corresponding value function $Q(s,a)$. $Q(s,a)$ represents the expected cumulative and discounted reward when the system starts from state $s$ and follows action $a$ and certain policy thereafter. $Q(s,a)$ for a discrete-time system is given by:
\begin{equation}
Q(s,a) = \mathbf{E} \Big[\sum_{k=0}^{\infty }\gamma^kr(k) \Big| s_{0},a_{0}\Big]
\end{equation}
where $r(t)$ is the reward rate and $\gamma$ is the discount rate in a discrete-time system.

In order to construct a DNN with a good accuracy, the offline phase needs to accumulate enough samples of $Q(s,a)$ value estimates and the corresponding state-action pairs $(s,a)$. It can be a model-based procedure or obtained from actual measurement data \cite{silver2016mastering}, in which the latter is the case for optimal DTRs in precision medicine.
This procedure includes simulating the control process, and obtaining the state transition profile and estimations for $Q(s,a)$ value, using an arbitrary but gradually refined policy.
The state transition profile is stored in an experience memory $D$ with capacity $N_{D}$.
According to the inventors of DRL \cite{mnih2015human}, the use of experience memory can smooth out learning and avoid oscillations or divergence in the parameters.
Based on the stored state transition profile and $Q(s,a)$ value estimates, the DNN is constructed with weight set $\theta$ trained using standard training algorithms such as backpropagation based stochastic gradient descent algorithms. The overall procedure is shown in the first part of Algorithm \ref{Alg:Sch}.

\begin{algorithm}[h]
	\caption{Illustration of the General DRL Framework}
	\label{Alg:Sch}
	\begin{algorithmic}[1]

		%\vspace{0.1in}
		%\REQUIRE $\mathcal{M}$, $\mathcal{N}$
		%\ENSURE $X = \langle x_{ij} \rangle$, $i \in \{ 1, ..., N \}$, $j \in \{ 1, ..., M \}$
		%\vspace{0.1in}
		%\hrule
		%\vspace{0.1in}
		\ENSURE {This is the \textbf{offline} part}
		\STATE Extract real data profiles using certain control policies and obtain the corresponding state transition profiles and $Q(s,a)$ value estimates;
		\STATE Store the state transition profiles and $Q(s,a)$ value estimates in experience memory $\mathcal{D}$ with capacity $N_\mathcal{D}$;
		\STATE Iterations may be needed in the above procedure;
		\STATE \textbf{Offline}: Pre-train a DNN with features $(s,a)$ and outcome $Q(s,a)$;
		\REQUIRE{This is the \textbf{online} part}
		%
		%\STATE $X = <x_{ij}> := 0, \forall i \in \{ 1, ..., N \}$, $j \in \{ 1, ..., M \}$
		%\newline
		%\STATE $X :=$ Even distribution of all threads in $\mathcal{T}$ to all servers in $\mathcal{S}$, i.e. $\forall i \in \{ 1, ..., T \}, x_{ij} := 1$ where $j: = (i$ mod $S)$ if $(i$ mod $S) \neq 0$, $j := S$ otherwise;
		%\newline
		\FOR{each execution sequence}
		\FOR{at each decision epoch $t_k$}
		\STATE With probability $\epsilon$ select a random action, otherwise $a_k =\ argmax_{a} Q(s_k, a)$, in which $Q(s_k, a)$ is derived (estimated) from DNN;
		\STATE Perform system control using the chosen action;
		\STATE Observe state transition at next decision epoch $t_{k+1}$ with new state $s_{k+1}$, receive reward $r_k(s_k,a_k)$ during time period $[t_k,t_{k+1})$;
		\STATE Store transition $\left(s_k, a_k, r_k, s_{k + 1}\right)$ in $\mathcal{D}$;
		\STATE Updating $Q(s_k,a_k)$ based on $r_k(s_k,a_k)$ and $\max_{a'}Q(s_{k+1},a')$ based on Q-learning updating rule;
		\ENDFOR
		\STATE Update DNN parameters $\theta$ using new Q-value estimates;
		\ENDFOR
	\end{algorithmic}
\end{algorithm}

For the online phase, the deep Q-learning technique is utilized based on the offline-trained DNN to select actions and update Q-value estimates. More specifically, at each decision epoch $t_k$ of an execution sequence, suppose the system under control is in the state $s_k$. The DRL agent performs inference using DNN to obtain the $Q(s_k,a)$ value estimate for each state-action pair $(s_k,a)$. Then according to the $\epsilon$-greedy policy, the action with the maximum $Q(s_k,a)$ value estimate is selected with a $1-\epsilon$ probability and a random action is selected with an $\epsilon$ probability. After choosing an action denoted by $a_k$, the DRL agent receives total reward $r_k(s_k,a_k)$ during $[t_k,t_{k+1})$ before the next decision epoch $t_{k+1}$, and this leads to Q-value updates. The reference work has proposed to utilize a duplicate DNN $\hat{Q}$ for Q-value estimate updating, in order to mitigate the potential oscillation of the inference results of the DNN\cite{lillicrap2015continuous}. At the end of the execution sequence, the DNN is updated by the DRL agent using the recently observed Q-value estimates in a mini-batch manner, and will be employed in the next execution sequence.
The overall procedure is shown in the second part of Algorithm 1.

As can be observed from the above procedure, the DRL framework is highly scalable for problems with a large state space, which is distinctive from the traditional reinforcement learning techniques. On the other hand, the DRL framework requires an enumerable action space due to the fact that at each decision epoch the DRL agent needs to enumerate all possible actions at the current state and perform inference using the DNN to derive the optimal $Q(s,a)$ value estimate (and corresponding optimal action). This implies that the action space in the general DRL framework, or for the specific optimal DTR problem, needs to be effectively reduced.

\subsection*{Developing a DRL Framework to Derive the Optimal DTR}

In this section, we present the developed DRL framework with a motivating DTR application using the database from CIBMTR registry on prevention and treatment of GVHD. There are two forms of GVHD: acute GVHD typically occurs within the first 6 months after the transplant and lasts for a short term if successfully treated; chronic GVHD may occur from shortly after the transplantation to a few years later, and often requires long-term treatment that can lead to long-term complications/mobidity.

Throughout this paper, we denote the time index $t=0$ for the time of transplantation, $t=1$ for 100 days, $t=2$ for 6 months, $t=3$ for 1 year, $t=4$ for 2 years, and $t=5$ for 4 years. We consider DTR within 4 years after the transplantation. In this paper, we adopt the DRL technique for three tasks of DTR: initial treatment before the transplantation including initial conditioning (chemotherapy to prevent relapse) and GVHD prophylaxis (to prevent GVHD), treatment of acute GVHD, and treatment of chronic GVHD. The initial preventive treatments take place at the time of transplantation $t=0$; the treatment of acute GVHD takes place at times $t=1$ (100 days) and $t=2$ (6 months); the treatment of chronic GVHD takes place at times $t=2$ (6 months) through $t=5$ (4 years).

As can be observed in Figure \ref{fig:figure1}, the proposed DRL framework for optimal DTR comprises two steps at each decision epoch/stage. The first step is to build a supervised learning network to predict the distribution of human experts' decisions on treatment actions. The second step is to estimate the value functions for treatment decisions with high probabilities (the actual implementation is also compatible with estimating value functions for all treatment options). In this way the proposed framework can both provide human experts' opinions and the data-driven comparisons of different strategies and recommendation for the optimal strategy, with relatively minor computational efforts. The proposed DRL framework is data-driven and scalable to the heterogeneous decision stages and high-dimensionality in patient features and treatment options. The DRL framework is adaptive, in that models in both steps will be updated when new data comes that correspond to new patients or treatment outcomes.

\subsubsection*{Retrieving the Target Cohort and Pre-Processing Data}
%The cohort of patients used for this analysis consists of 6021 patients diagnosed with Acute Myeloid Leukemia (AML) and received transplant between 1995 and 2007.  CIBMTR collect longitudinal data from transplant through standard forms.The time stamps we collected a sub set of these follow-up information in 100 days, 6 months, 1 year,2 year and 4 year with the limited time and research resources.

\begin{figure}
\caption{Histogram of Patient Ages in the Data Set of Interests}
\label{age}
\includegraphics[scale=0.5]{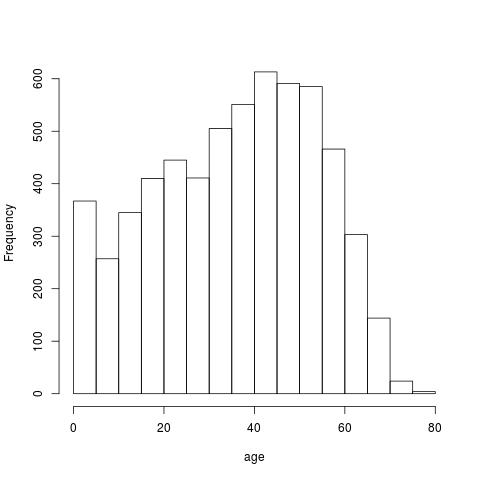}.
\end{figure}

The cohort of patients used for this analysis consists of 6,021 patients diagnosed with Acute Myeloid Leukemia (AML) who have undergone HCT between 1995 and 2007. Due to the discrete data collection scheme, we have higher quality data on the onsets of GVHD conditions and the subsequent treatment decisions in a discrete-time frame indicating the occurrence between two follow-up times. The exact date and sequence of treatment decisions between two periods of time are missing or not recorded to a greater extent. In this work, the state and action are considered to be the state and action taken at the time each form was recorded. We consider relapse and death as terminal states and occurrences of acute or chronic GVHD as transient states. We consider baseline features of patients and donors that have been shown to affect GVHD and survival rates in clinical studies, including patient's age, gender, and co-morbidity information (including diabetes, seizure, hypertension, etc.). It also includes important patient and donor relationship to patient and matching information, as well as donor's gender.  This cohort includes both pediatric and adult patients. We include the histogram of age in Figure \ref{age}. And the Human Leukocyte Antigen (HLA) matching results of patients are presented in Table \ref{match}.

\begin{table}
\caption{Matching Information of Patients and Donors in the Data Set of Interests}
\label{match}
\begin{tabular}{cccccc}
\hline
Identical Sibling&Other Relative&URD Well Matched & URD Partially Matched & URD Mismatched & Other\\
\hline
3877 & 451& 686& 433& 173&401\\
\hline
\end{tabular}
\end{table}

%These variables have relatively less missing, and missing is considered as one category if presented.

%We first used autoencoder to conduct dimension reduction on these features....

\subsection*{Building a Deep Neural Network to Predict Expert Treatments}
\begin{figure}[t]
  \centering
  \includegraphics[width=0.8\columnwidth]{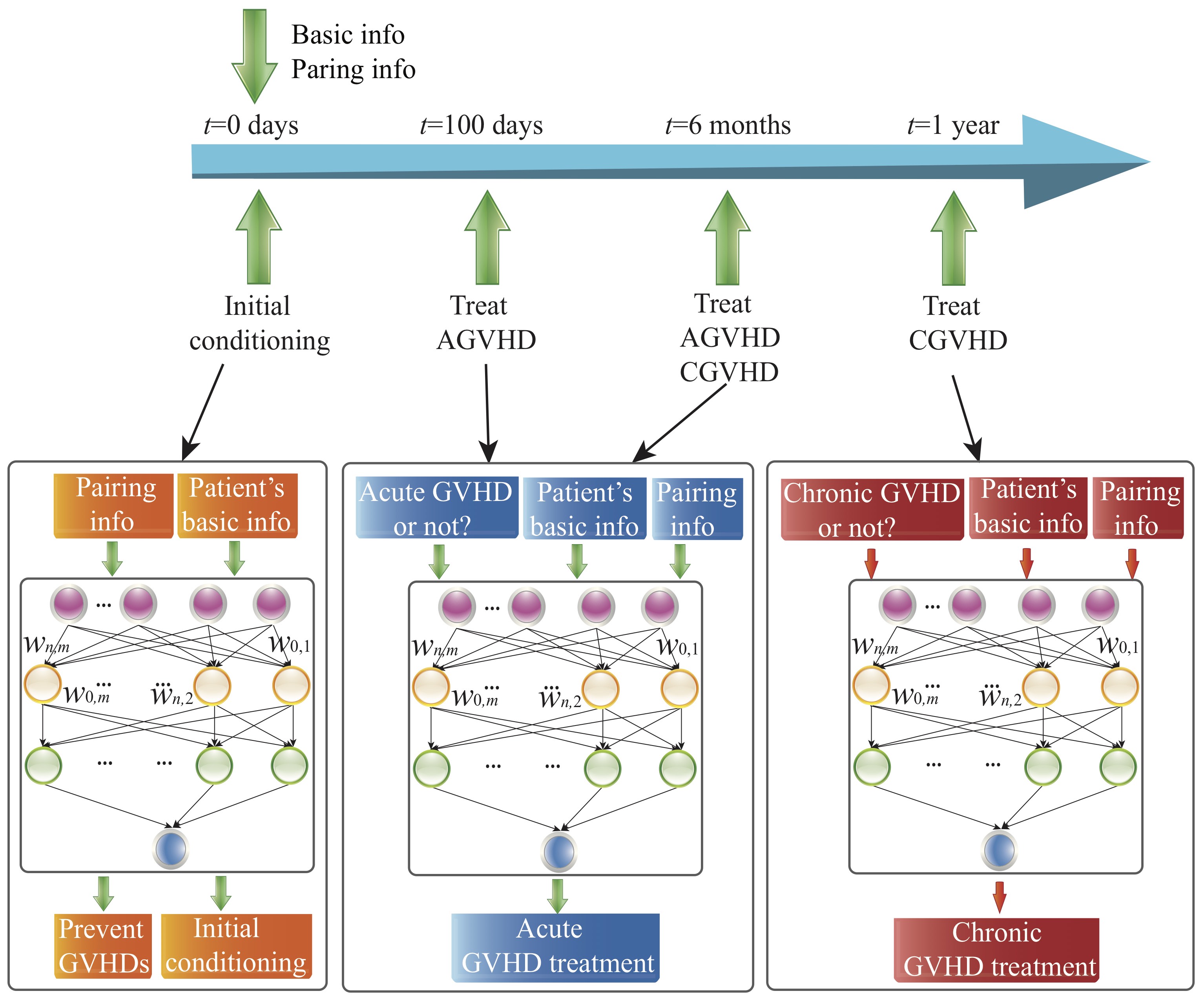}
  \caption{The proposed DRL framework for prevention and treatment of GVHD, as well as initial conditioning.}
  \label{fig:figure1}
\end{figure}

As shown in Figure \ref{fig:figure1}, the first step at each decision epoch is to build a supervised learning network to predict the distribution of human experts' decisions on treatment actions. The prediction networks are illustrated in Figure \ref{fig:figure1}. For the initial treatment before the transplantation, the input features (the state space) include the union of the basic information of patients (e.g., age, gender, and cormorbilities, etc.) and the HLA matching information between the patient and the donor. The output label (action) is the combination of medicines to be utilized for the initial treatment which include the initial conditioning to avoid disease relapse and the GVHD prophylaxis to prevent GVHD.

For the treatment of acute GVHD at time stamps $t=1$ and $t=2$, the input features (the state space) include both the basic information of patients and the pairing conditions, as well as whether the patient has acute GVHD at that specific time stamp. The output label (action) is the combination of medicines to be utilized for the treatment of acute GVHD. Similar input features and actions also apply to the treatment of chronic GVHDs from $t=2$ through $t=5$.

To reduce the high dimensionality in the action space, we encode the actions using all the medicine combinations that have been already utilized by doctors. We adopt an effective encoding scheme of the state space to reduce the state space to a large extent, thereby accelerating the convergence speed and mitigating potential overfitting issues. For enhancing the accuracy, separate multi-layer deep neural networks (instead of a single integrated network) are trained offline for the initial conditioning, prevention of GVHDs, treatment of acute and chronic GVHDs.

In this step, we adopt the multiple layer, fully-connected neural network as our deep neural network.
The network architecture consists of four layers: the input layer, two hidden layers and the output layer. The dimension of the input layer is 9 and the two hidden layers have 16 and 32 neurons, respectively. The output dimension is 145 for initial conditioning and 127 for GVHD prophylaxis. The output dimensions for treating acute and chronic GVHDs are 283 and 271, respectively. We use the Adam Optimizer to train the network, and the learning rate $\eta$ is set to be $10^{-4}$ \cite{DBLP:journals/corr/KingmaB14}.

\subsubsection*{Estimating Value Function for Top Expert Choices and Making Recommendations}

As shown in Figure \ref{fig:figure1}, the second step is to estimate the value function for expert actions with highest probabilities and make recommendations among treatment options. Our recommender only evaluates value function for actions with highest probabilities, since actions with small probability have too small number of samples in the observational medical datasets to arrive at a general conclusion. This restriction also reduces the computational complexity. The reward/outcome of major interests is the \emph{relapse-free survival time} after the transplantation, denoted as $T_i$.
Let $\vec{a}$ denote the vector of actions at all stages, $\vec{\pi}$ denote the rules of decision sequences (i.e., policies), which represent the mapping from the currently observed state to action at each stage.
The value function of a policy $\vec{\pi}$ is $V(\vec{\pi})=E(T_i| \vec{a}\in \vec{\pi} (s) )$. The objective is to maximize $V(\vec{\pi})$ and the so-called Q-function is the expected reward if a subject (patient) is assigned to the optimal treatment in all the future stages, and can be estimated through \emph{Dynamic Programming} following the ideas from Q-learning \cite{watkins1992q}.
The learning algorithm will be tailored for the specific \emph{Censoring Scheme} of the data set. Denote $M_i$ as the indicator of whether patient $i$ is censored ($M_i=1$ if death or relapse is observed), and $C_i$ as the last observation time of patient $i$. Denote $D_{t,i}$ as the indicator that death or relapse is observed within the time period $t-1$ to $t$. For time $t$ and patient $i$, denote the indicator of observed terminal events at time $t$ as $M_{t,i}= \mathbb{I}(D_{1,i}=0,\dots,D_{t-1,i}=0,D_{t,i}=1)$, where $\mathbb{I}(\cdot)$ is the indicator function.
The general Q-learning uses a backward induction procedure across time stamps (decision stages).
At stage $t$, each valid training sample (patient) $i$ needs to satisfy $C_i>t$, and requires that action $a_t$ to be observed. For patients with $M_{t+1,i}=1$, we use their observed $T_i$ as the outcome. For patients with $C_i > t+1$, or $C_i=t+1, M_i=0$, we use the estimated Q-function for the future stage as the outcome. In other words, we impute patients who have survived beyond time stamp $t+1$ using their optimal future value estimation, regardless of censoring.

Besides the value function of relapse-free survival time, we also propose to use an alternative discretized value function as shown in the following. For each patient $i$, let $t_i$ denote the time when he/she enters the terminal state (death, relapse, or relapse-free survival after 4 years) or when his/hers data get lost. The delayed reward (outcome) of patient $i$ at time $t_i$ can be classified into the following categories:
\begin{enumerate}
\item Relapse-free and GVHD-free survival.
\item Survival with acute or chronic GVHD.
\item Relapse of the leukemia disease.
\item Death.
\item Data loss.
\end{enumerate}
We assign different delayed rewards/outcomes for the five cases. For relapse-free and GVHD-free survival in 4 years, the highest reward (1) is achieved. Survival with acute or chronic GVHD receives a slightly degraded reward (0.8). Relapsed patients receive a significantly degraded reward (0.2). Death receives zero. This reward can be viewed as a heuristic 4-year-survival probability adjusted for the quality-of-life. The missing data problem caused by the lost of follow-up is solved by the imputation method as discussed above.

In order to accommodate high dimensionality in state and action spaces, the recent DRL literature implemented Q-learning with deep neural networks to approximate the Q-function, which is named as the \emph{Deep Q-Network}. In this problem of interest, three separate deep Q-networks are developed for DTRs of initial conditioning (chemotherapy and prevention of GVHDs), treatment of acute and chronic GVHDs. For the inputs of deep Q-networks at time $t$, the corresponding input states described in the previous section (predicting human experts' decisions) serve as states, and the predicted human experts' decisions serve as actions. Effective encoding scheme is utilized to reduce the input state space. The output prediction is the expected value/return starting at this state and taking the corresponding action. Multi-layer deep neural networks are constructed to achieve this goal, and only those patients whose data are available at each time $t$ are utilized to train the deep Q-networks.

In the deep Q-network, we use a replay buffer to store the dataset\cite{lillicrap2015continuous}. The replay buffer is a finite sized cache which can store the sampled transition tuples $(s_t,a_t,r_t,s_{t+1})$, and it discards the oldest samples when it is full. The reply buffer allows the algorithm to benefit from learning across a set of uncorrelated transitions. Direct implementation of deep Q-learning may cause the network to be unstable during the training process. As a result, we adopt the target network introduced in reference\cite{lillicrap2015continuous}. The target network is a copy of the Q-value network and is used to perform inference of $Q(s_{k+1},a')$. The weights of the target network are updated by slowly tracking of the updated parameters in the Q-value network: $\theta' \gets \tau\theta+(1-\tau)\theta'$ with $\tau \ll 1$. This constraint can significantly improve the stability of learning.

Similar to the first step, a four layer fully-connected neural network architecture is adopted in the DRL network for acute and chronic GVHD treatments. It consists of the input layer, two hidden layers and the output layer. The input dimensions for treating both acute and chronic GVHDs are 8. The output dimensions for treating acute and chronic GVHDs are 283 and 271, respectively. The numbers of neurons in the two hidden layers are 32 and 64 for both acute and chronic GVHD treatments. The learning rate $\eta$ is set to be $10^{-3}$. The target network updating parameter $\tau$ is set as 0.01, and the discount rate of reward $\gamma$ is set as 0.99. The size of replay buffer is 20000.

\bibliography{star}

\section*{Acknowledgement}
Our thanks to the CIBMTR to provide the dataset. The CIBMTR is supported primarily by grants  NCI/NHLBI/NIAID 5U24CA076518, and NCI/NHLBI 5U10HL069294.

\section*{Competing financial interests}
The author(s) declare no competing financial interests.

\section*{Author contribution statement}
Ying Liu initiated the idea after discussion with Yanzhi Wang. Brent Logan helped identified the clinical data set and give guidance on the formulation of dynamic treatment regimes and data cleaning. Ying Liu cleaned the data and aligned the data set in terms of state and actions for reinforcement learning framework. Ning Liu mainly designed the structure of the neural networks and deep Q networks and conducted the analysis. Zhiyuan Xu and Jian Tang also helped with the data analysis. Ying Liu, Yanzhi Wang and  Ning Liu write different part of the manuscript.

\section*{Availability of materials and data}
The authors are willing to share the cleaned data and code with the Editorial Board Members and referees upon request. The cleaned data are identified human subject information summarized in forms of state and actions at each decision point, additional steps of IRB approval maybe needed in accordance with CIBMTR registry policy.

\end{document}